\documentclass[acmsmall]{acmart}

\AtBeginDocument{%
  \providecommand\BibTeX{{%
    \normalfont B\kern-0.5em{\scshape i\kern-0.25em b}\kern-0.8em\TeX}}}

\setcopyright{acmcopyright}
\copyrightyear{2020}
\acmYear{2020}

\acmJournal{TOMM}

\usepackage{multirow}
\usepackage{xcolor}
\definecolor{newcolor}{rgb}{.8,.349,.1}
\newcommand{\bc}[1]{\textcolor{blue}{#1}}
\newcommand{\rc}[1]{\textcolor{red}{#1}}
\newcommand{\gc}[1]{\textcolor{green}{#1}}
\usepackage{amssymb}
\usepackage{pifont}

\begin{document}

\title{Exploring Image Enhancement for Salient Object Detection in Low Light Images}


\author{Xin Xu}
\authornotemark[1]
\affiliation{%
  \institution{Wuhan University of Science and Technology}
  \city{Wuhan}
  \country{China}}
\email{xuxin@wust.edu.cn}

\author{Shiqin Wang}
\affiliation{%
  \institution{Wuhan University of Science and Technology}
  \city{Wuhan}
  \country{China}
}

\author{Zheng Wang}
\affiliation{%
 \institution{National Institute of Informatics}
 \city{Tokyo}
 \country{Japan}}
\email{wangz@nii.ac.jp}

\author{Xiaolong Zhang}
\affiliation{%
  \institution{Wuhan University of Science and Technology}
  \city{Wuhan}
  \country{China}}
\email{xiaolong.zhang@wust.edu.cn}

\author{Ruimin Hu}
\authornotemark[1]
\affiliation{%
  \institution{Wuhan University}
  \city{Wuhan}
  \country{China}}
\email{hrm1964@163.com}

\renewcommand{\shortauthors}{Xin Xu and et al.}

\begin{abstract}
  Low light images captured in a non-uniform illumination environment usually are degraded with the scene depth and the corresponding environment lights. This degradation results in severe object information loss in the degraded image modality, which makes the salient object detection more challenging due to low contrast property and artificial light influence. However, existing salient object detection models are developed based on the assumption that the images are captured under a sufficient brightness environment, which is impractical in real-world scenarios. In this work, we propose an image enhancement approach to facilitate the salient object detection in low light images. The proposed model directly embeds the physical lighting model into the deep neural network to describe the degradation of low light images, in which the environment light is treated as a point-wise variate and changes with local content. Moreover, a Non-Local-Block Layer is utilized to capture the difference of local content of an object against its local neighborhood favoring regions. To quantitative evaluation, we construct a low light Images dataset with pixel-level human-labeled ground-truth annotations and report promising results on four public datasets and our benchmark dataset.
\end{abstract}



\maketitle

\begin{figure}[h]
\centering
\includegraphics[width=0.9\columnwidth]{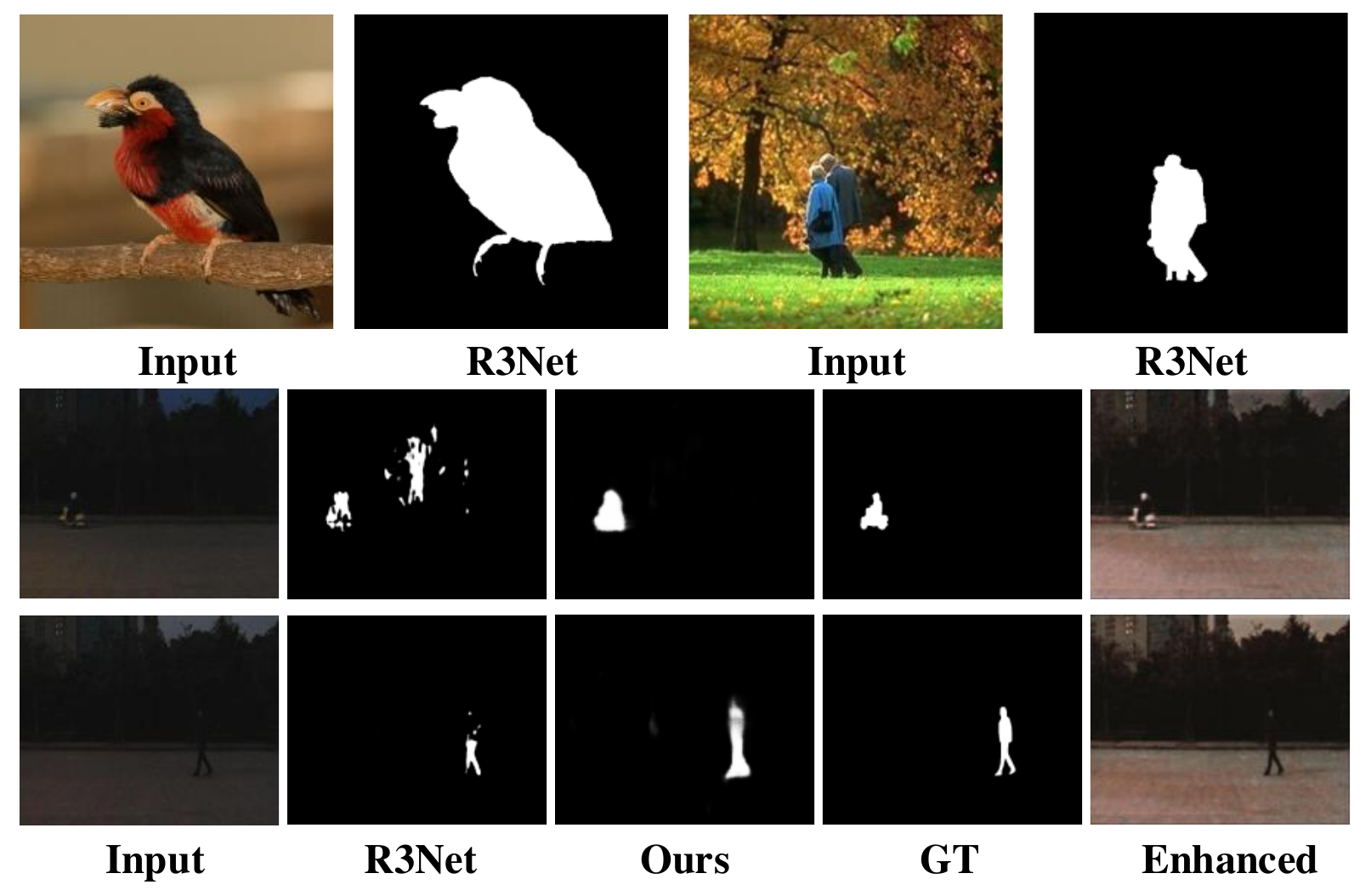}
\caption{{\bf Examples and their SOD results.} The first row shows two examples in the general SOD task. The inputs are respectively the images from the DUT-OMRON~\cite{yang2013saliency} and PASCAL-S~\cite{li2014secrets} datasets. Their corresponding results are generated by R3Net~\cite{deng2018r3net}. The SOD performances are perfect for these two images with sufficient brightness. The next two rows show the results of the low light images. From left to right are respectively the input images, the results by R3Net, the results by our approach, the ground truths, and the enhanced results of our approach. R3Net performs not so good, while our approach achieves considerable results.}
\label{fig:problem}
\end{figure}

\section{Introduction}
Salient Object Detection (SOD) aims at localizing and segmenting the most conspicuous objects or regions in an image. As a pre-processing step in computer vision, SOD is of interest to urban surveillance and facilitates a wide range of visual applications, \textit{e.g.} object re-targeting~\cite{li2018depth,wu2018deep,wu2019few,wang2012event,wu2019}, semantic segmentation~\cite{zhang2019synthesizing,zhang2015spatiochromatic}, image synthesis~\cite{zhan2018verisimilar,chen2019quality,wang2020}, visual tracking~\cite{feng2019dynamic,wang2012movie2comics,wang2015visual}, image retrieval~\cite{wei2019saliency,wu20193,9055370,wang2017effective,wu2018cycle}, and \textit{etc}.

Current SOD methods primarily utilize global and local features to locate salient objects on existing SOD datasets. Images in these datasets are usually captured in the environment with sufficient brightness. However, the effectiveness of current SOD methods in low light images is still limited. Images captured in low illumination conditions usually exhibit low contrast and low illumination properties. These properties cause severe object information loss in dark regions, where the salient object is hard to detect. As shown in the second and third columns of Fig.~\ref{fig:problem}, the results of R3Net lose detail information of salient objects and tend to contain non-saliency backgrounds in the degraded low light images. The reason mainly attributes to the fact that the environment light in low light image modality primarily consists of artificial light. Because of the influence of the artificial light, the environment light is ever-changing at different locals of the image. Thus, the environment light, working as noise, will degrade the image capturing process.

Different from existing SOD methods that conduct SOD directly on original degraded images, we eliminate the effect of low illumination by explicitly modeling the physical lighting of the environment for image enhancement. The detail information of the salient object can be retained to improve the SOD performance. To achieve this goal, it is natural to enhance low light image first. However, existing low light image enhancement mainly focus on improving subjective visual quality, rather than facilitating subsequent high-level SOD task. To alleviate such a problem, we first embeds the physical lighting model into the deep neural network to describe the degradation of low light images, in which the environment light is treated as a point-wise variate and changes with local content. Then a Non-Local-Block Layer is utilized to extract non-local features of salient objects. Moreover, a low light image dataset is built to evaluate the performance of SOD.

In summary, the main contribution of our work is threefold:

\begin{itemize}
\item We build a low light image dataset for the SOD community. Based on this dataset, we verify that low illumination can reduce the performance of SOD.
\item We explore image enhancement for low illumination SOD. The effect of low illumination can be eliminated by explicitly modeling the physical lighting of the environment for image enhancement, and the detail information of the salient object can be retained to improve the SOD performance.
\item To account for the non-uniform environment light, the physical lighting of low light images is analyzed to build the degradation model, where the environment light is treated as a point-wise variate and changes with the local light source. Moreover, a Non-Local-Block Layer is utilized to capture the difference of local content of an object against its local neighborhood favoring regions.
\end{itemize}

\section{Related works}
SOD has achieved remarkable progress in recent years and is a hot topic in both academic and industrial communities. The main difficulty in SOD is how to separate salient object from its surroundings to resist the interference caused by variations in viewpoint, background, resolution, illumination, \textit{etc}. Inspired by current low-light image enhancement approaches, this paper focuses on the low illumination issue in the SOD task.

\subsection{Salient Object Detection}
Traditional SOD models locate conspicuous image regions by computing the difference with their surroundings and primarily rely on hand-crafted features. These models do not require the training process, and extract saliency feature from color~\cite{deng2010generalized}, contrast~\cite{li2015low}, contour~\cite{zhang2015spatiochromatic}, objectness~\cite{chang2011fusing}, focusness~\cite{jiang2013salient}, and backgroundness~\cite{qin2015saliency}. In recent years, deep learning based SOD models extract high level features in a data-driven way and have demonstrated their superior performance. These high level features can be broadly divided into three categories: global feature, local feature, and global \& local feature.

Li et al.~\cite{li2016deepsaliency} proposed the multi-task (MT) neural network which uses convolution to extract global features. Zeng et al.~\cite{zeng2018learning} formulated zero-shot learning to promote saliency detectors (LPS) to embedded DNN as a global feature extractor into an image-specific classifier. While global features can only roughly determine the location of salient object with incomplete information, Li and Yu~\cite{Li2015Visual} proposed the multi-scale deep features (MSDF) neural network which decomposes input images into a set of non-overlapping blocks and then puts them into the three-scale neural networks to learn local features. Deng et al.~\cite{deng2018r3net} proposed a recurrent residual refinement network (R3Net) to learn local residual between the non-salient regions from intermediate prediction and saliency details from the ground truth. Similarly, Qin et al.~\cite{qin2019basnet} proposed a Boundary-aware salient object detection network (BASNet) to conduct a coarse prediction with the residual refinement to hierarchically extract local features.

However, multiple levels of local convolutional blur the object boundaries, and high level features from the output of the last layer are spatially coarse to perform the SOD tasks. Recent work attempted to combine the information from both global and local features. Yang et al.~\cite{yang2013saliency} utilized graph-based manifold ranking (MR) to evaluate the similarity of local image pixels with global foreground cues or background cues. Luo et al.~\cite{luo2017non} utilized the non-local deep features (NLDF) to connect each local feature, and fused with the global features to output the saliency map. Hou et al.~\cite{hou2017deeply} proposed deeply supervised SOD with short connections (DSSC) from deep-side outputs with global location information to shallow ones with local fine details.

To the best of our knowledge, quite a few works attempt to address the issue of low illumination for SOD. Due to the effect of low illumination, images captured in a non-uniform illumination environment usually are degraded with the scene depth and the corresponding environment lights. This degradation results in severe object information loss, which makes the SOD more challenging. In~\cite{xu2018extended}, we previously proposed to extract non-local features for SOD in low light images. However, conducting low illumination SOD directly on original degraded images may be non-optimal. In this work, we focus on the problem of low light image enhancement for SOD.

\subsection{Low Light Image Enhancement}
In the past decades, various enhancement techniques have been developed to improve the quality of low light images. Histogram equalization (HE) is a widely utilized approach due to its simplicity. Global HE approach~\cite{tomasi1998bilateral} balances the histogram of the entire image, and is suitable for lightening the overall low light images. However, global HE enables the gray levels with high-frequency to dominate the other low-frequency gray levels and may degrade sharpness at the boundary. To tackle this problem, local HE approach~\cite{lee2013contrast} conducts the calculation inside a sliding window over the low light image. However, local HE may cause over-enhancement in bright regions.

Another choice is based on Retinex~\cite{land1986alternative} and multi-scale Retinex (MSR)~\cite{jobson1997multiscale} which assume that an image can be composed of scene reflection and illumination. Fu et al.~\cite{fu2016weighted} proposed a weighted variational model to adjust the regularization terms by the fusion of multiple derivations of illumination map. However, this method ignores the structure of illumination and may lose realistic in rich textures regions. Guo et al.~\cite{guo2017lime} proposed a low-light image enhancement (LIME) approach to estimate the illumination of each pixel in RGB channels which is refined by structure prior. Retinex based approaches are based on the Lambertian scene assumption and require illumination environment should be piece-wise smooth. However, low light images captured in low illumination environments usually contain regions with rapidly changing illumination due to artificial light interference, which may cause Halo effects in these regions.

To tackle the non-uniform illumination issue, an alternative way is to analyze the physical lighting of low light images. Dong et al.~\cite{dong2011fast} assumed that the inverted low light images are similar to images captured in hazy conditions, and applied dark channel prior to analysis the image degradation. However, the image degradation model for the haze environment is inadequate to reflect the globally physical lighting and cause potential information loss in the dark regions of low light images. Ying et al.~\cite{ying2017new} and Ren et al.~\cite{ren2018lecarm} utilized the illumination estimation technique to obtain the exposure ratio map and incorporated a camera response model to adjust image pixel according to the exposure ratio to solve lightness distortion.

Above conventional low light image enhancement approaches rely heavily on the parameter tuning to improve the subjective and objective quality of low light images. Recently, deep learning based methods have been widely investigated, enhancing low light images directly in a data-driven way. Wei et al.~\cite{wei2018deep} constructed a Retinex based image decomposition network (RetinexNet) to learn the end-to-end mapping between low illumination and normal light image pairs. Wang et al. ~\cite{wang2018gladnet} proposed a GLobal illumination-Aware and Detail-preserving Network (GLADNet), including a global illumination estimation step and a detail reconstruction step. However, existing low light image enhancement mainly focus on improving subjective visual quality, rather than facilitating subsequent high-level SOD task.

\section{Proposed Method}
The framework of our method consists of two sub-networks, i.e., Physical-based Image Enhancement (PIE) subnet and Non-Local-based Detection (NLD) Subnet. PIE enhances the image contrast by exploiting the relation among the atmosphere light $A(z)$ and the transmission map $t(z)$. NLD detects salient object from the enhanced images $J(z)$. Fig.~\ref{fig:framework} shows the framework of our method. We explain these sub-networks in detail as follows.

\subsection{PIE}
Floating particles in the atmosphere greatly scatter the environment light in the nighttime scene, resulting in degradation in the image quality. This degradation causes severe object information loss in dark regions and in turn affects the performance of SOD. PIE aims at generating a better image $J$ from the given low light image $I$ then benefiting the SOD task.

Inspired by the dehazing method DCPDN~\cite{zhang2018densely} which is based on the atmospheric scattering model, we proposed the PIE for low light image enhancement. Although the atmospheric scattering model is utilized in the DCPDN for image dehazing, it is also capable of analyzing physical lighting of low light images, because of the existence of atmospheric particles in the nighttime scene. Therefore, following the DCPDN, the PIE also consists of four key modules, including a U-Net, an encoder-decoder network, an atmospheric scattering model, and a joint discriminator. However, different from the constant environment light in the typical hazy model, low light images are usually taken in non-uniform environmental light. The atmospheric light is treated as a point-wise random variate in PIE rather than a constant in DCPDN, to follow the rules of nighttime light.

U-Net is exploited to predict the atmospheric light $A(z)$. The encoder-decoder network is used to estimate the transmission map $t(z)$. Combining the results of $A(z)$ and $t(z)$, the atmospheric scattering model generates the enhanced image $J(z)$. Since the enhanced image and its corresponding transmission map should have strong structural relationship, $t(z)$ and $J(z)$ are concatenated together and the joint discriminator is used to distinguish whether a pair of estimated $t(z)$ and $J(z)$ is a real or fake pair.

\subsubsection{U-Net.} We used an 8-block U-Net structure ~\cite{Ronneberger2015U} to estimate the atmospheric light. The U-net can preserve rich texture information and has achieved spectacular performance on image enhancement~\cite{lv2019attention,chen2018deep,huang2018range,jiang2019enlightengan}. Another advantage of using the U-Net lies in its efficient GPU consumption. The U-Net consists of an encoder and a decoder. They are connected like a `U' shape. The encoder is composed of four Conv-BN-Relu blocks, while the decoder is composed of symmetric Dconv-BN-Relu block (Con: Convolution, BN: Batch-normalization, and Dconv: Deconvolution).

As we know, images captured in a low illumination environment are degraded with the corresponding environment lights. It is impossible to describe the changes in the incident light for each image pixel at the same level. Different from the constant environment light in the typical hazy model~\cite{zhang2018densely}, low light images are usually taken in non-uniform environmental light. Therefore, we treat the atmospheric light $A(z)$ as a point-wise variate, changing with the local scene light source. Thus, we synthesize the training samples for U-net, where $A(z)$ is randomly valued to generate the corresponding atmospheric light maps. It can be formulated as:

\begin{equation}
A(z)=1-\alpha*uniform(0,1),
\end{equation}
where $uniform(0,1)$ randomly generates real numbers between 0 and 1. To simplify our method, we set $\alpha = 0.5$ in this paper.

\subsubsection{Encoder-decoder Network.}
We used an encoder-decoder network to estimate the transmission map $t(z)$. The encoder-decoder architecture has achieved spectacular performance on image dehazing~\cite{zhang2018} and image enhancement~\cite{wei2018deep,ren2019}. The encoder-decoder network can keep the structural information of the object and produce the high-resolution feature map from low-resolution saliency map. The encoder is composed of the first Conv layer and the first three Dense-Blocks with their corresponding down-sampling operations Transition-Blocks from a pre-trained dense-net121. The decoder consists of five dense blocks with the refined up-sampling Transition-Blocks. The function of the encoder is to leverage the pre-defined weights of the dense-net~\cite{huang2017densely} and the function of the decoder is to reconstruct the transmission map into the original resolution.

\subsubsection{Atmospheric Scattering Model.}
After estimating the atmospheric light $A(z)$ and transmission map $t(z)$, the target image $J(z)$ can be estimated via the atmospheric scattering model. The atmospheric scattering model is derived from McCartney's scattering theory, which assumes the existence of atmospheric particles and has been put into practice in haze removal~\cite{ancuti2013single}~\cite{zhang2016nighttime}. The atmospheric scattering model is also suitable for low light image enhancement because there are similarities between low light images and hazy images. The scattering particles exist everywhere, even on clear sunny days~\cite{he2010single}, the scattering phenomenon caused by which is a cue to the aerial perspective~\cite{preetham1999practical}. Therefore, the existence of light scattering is necessary for low light images. Accordingly, the atmospheric scattering model for low light images can be composed of two items: the direct attenuation term and the scattering light term. The former represents the object reflects light which is not scattered by the scattering particles. While the later is a part of the scattered environment light that reaches the camera. The atmospheric scattering model for low light image can be mathematically expressed as:

\begin{equation}
I(z)=J(z) t(z)+A(z)(1-t(z)),
\end{equation}
where $J$ is the enhanced target image, $I$ is the observed low light image, $z$ is the location of the pixel. Different from the constant environment light in the typical hazy model, $A$ is a point-wise variate and changes with the local scene light source.

\subsubsection{Joint Discriminator Learning.}
According to Zhang et al.~\cite{zhang2018densely}, the structural information between the transmission map $t(z)$ and the enhanced image $J(z)$ is highly correlated. Therefore, we use the joint discriminator learning to refine the enhanced image $J(z)$. The joint discriminator learning aims to make sure that the estimated transmission map $t(z)$ and the enhanced image $J(z)$ are indistinguishable from their corresponding ground truths, respectively. It is formulated as:

\begin{equation}
\begin{split}
\min _{G_{t}, G_{d}} \max _{D_{joint}}&\mathrm{E}_{I \sim Pdata(I)}\left[\log \left(1-D_{joint}\left(G_{t}(I)\right)\right)\right]+\\
&\mathrm{E}_{I \sim Pdata(I)}\left[\log \left(1-D_{joint}\left(G_{d}(I)\right)\right)\right]+\\
&\mathrm{E}_{t, J \sim Pdata(t,J)}\left[\log D_{joint}(t, J)\right)],
\end{split}
\end{equation}
where $G_{t}$ and $G_{d}$ denote the networks generating the transmission map and the enhanced result, respectively. The joint discriminator learning process exploits the structural correlation between the transmission map and the enhanced image.

\begin{figure}[h]
\centering
\includegraphics[width=1\textwidth]{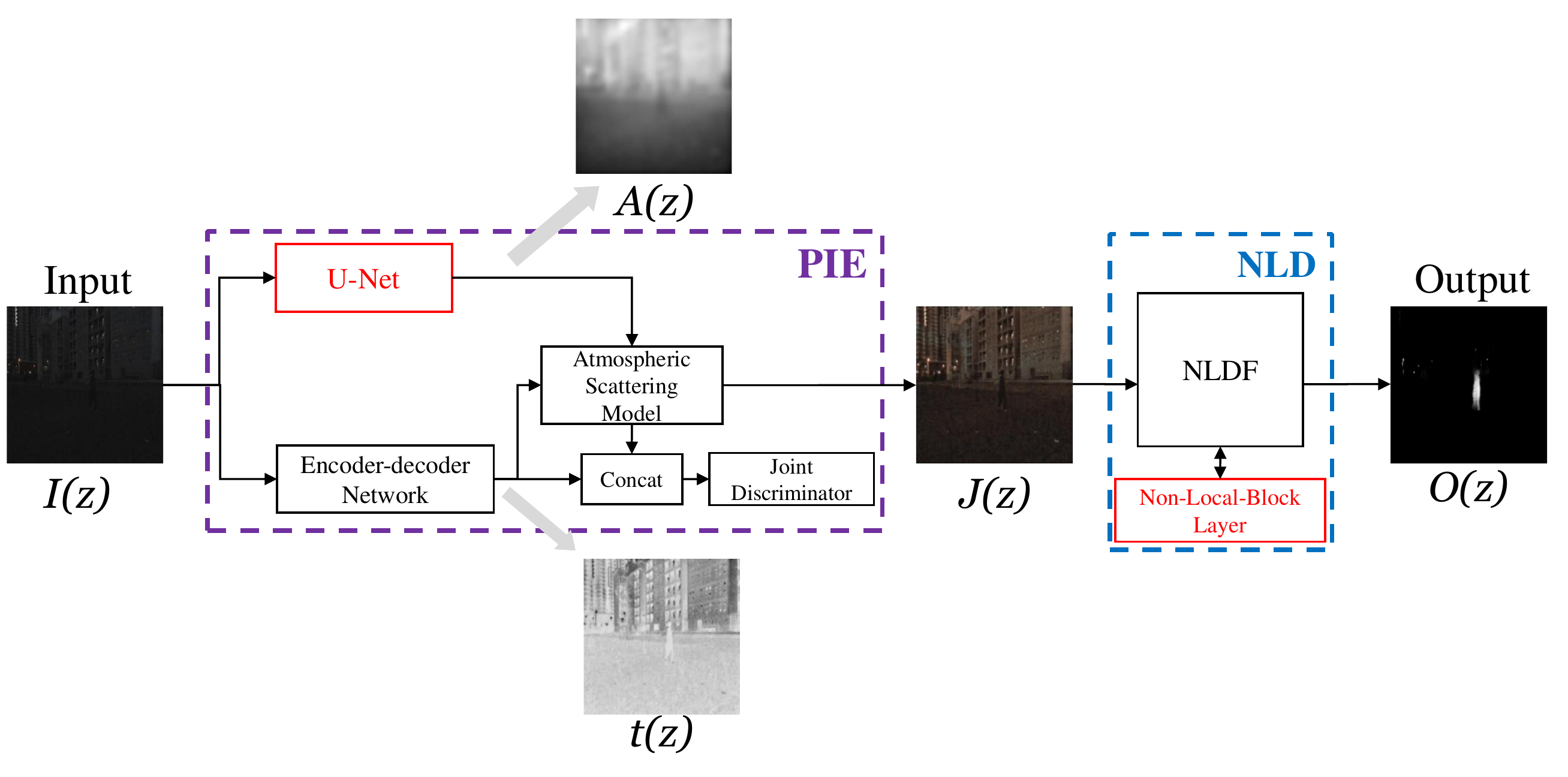} 
\caption{{\bf An overview of the proposed method.} The framework of our method consists of two sub-networks: Physical-based Image Enhancement (PIE) subnet and Non-local-based Detection (NLD) subnet. PIE attempts to generate a better image $J(z)$ from the given low light image $I(z)$ then benefit the SOD task. NLD aims to generate saliency map $O(z)$ from the enhanced image $J(z)$ by learning discriminant saliency features from the nighttime scene. For PIE, we treat the atmospheric light $A(z)$ as a point-wise random variate rather than a constant, to follow the rules of nighttime light. For NLD, we reform the NLDF~\cite{luo2017non} by adding the Non-Local-Block Layer, to provide a robust representation of saliency information towards low light images captured in non-uniform artificial light.
}
\label{fig:framework}
\end{figure}

\begin{figure}[h]
\centering
\includegraphics[width=0.8\columnwidth]{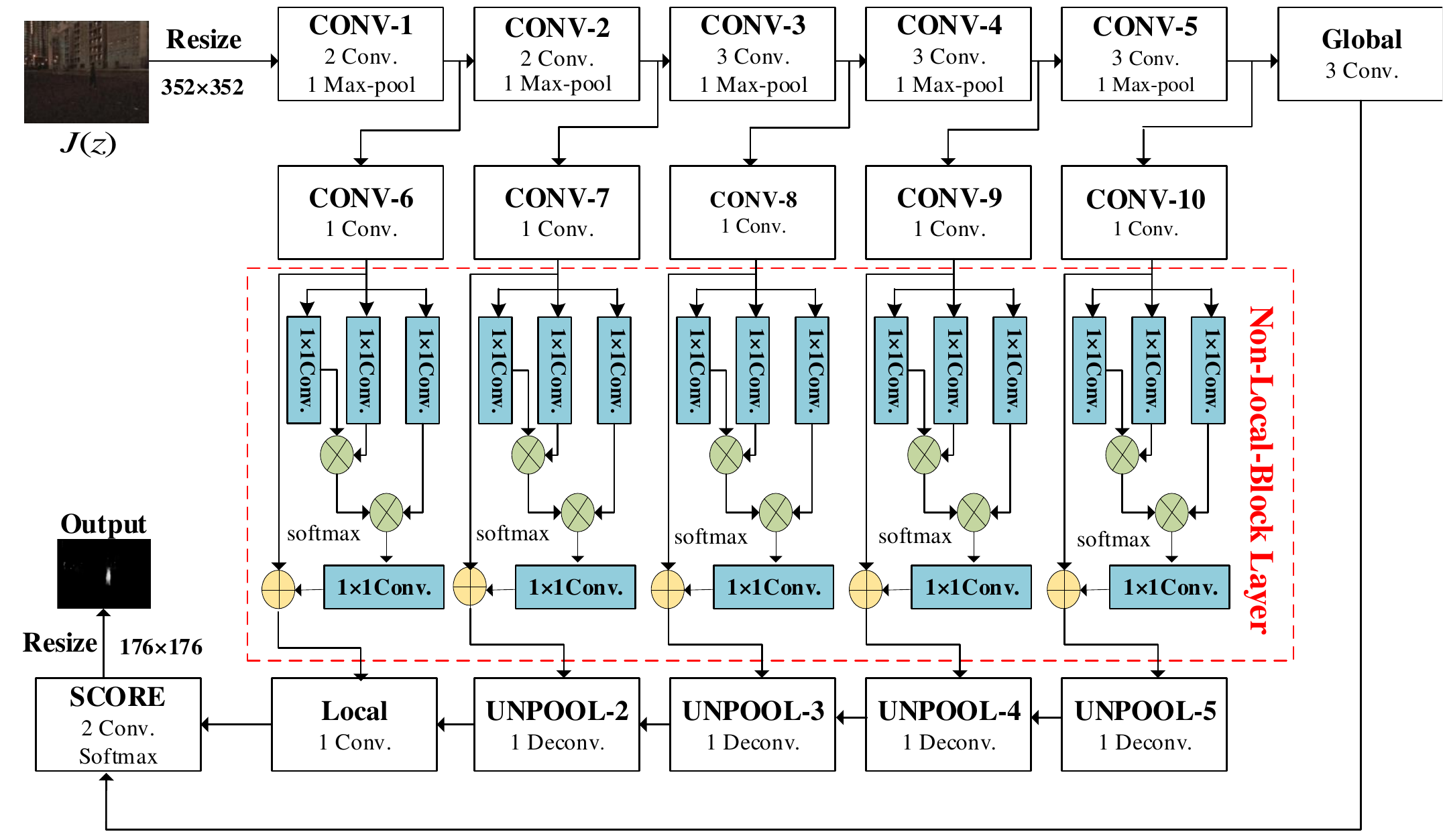}
\caption{{\bf Architecture of Non-Local-based Detection (NLD) subnet for salient object detection.} $J(z)$ is the output of the former PIE and the input of NLD. The red region indicates the proposed Non-Local-Block Layer.}
\label{fig:NLD}
\end{figure}

\begin{figure}[h]
\centering
\includegraphics[width=0.6\columnwidth]{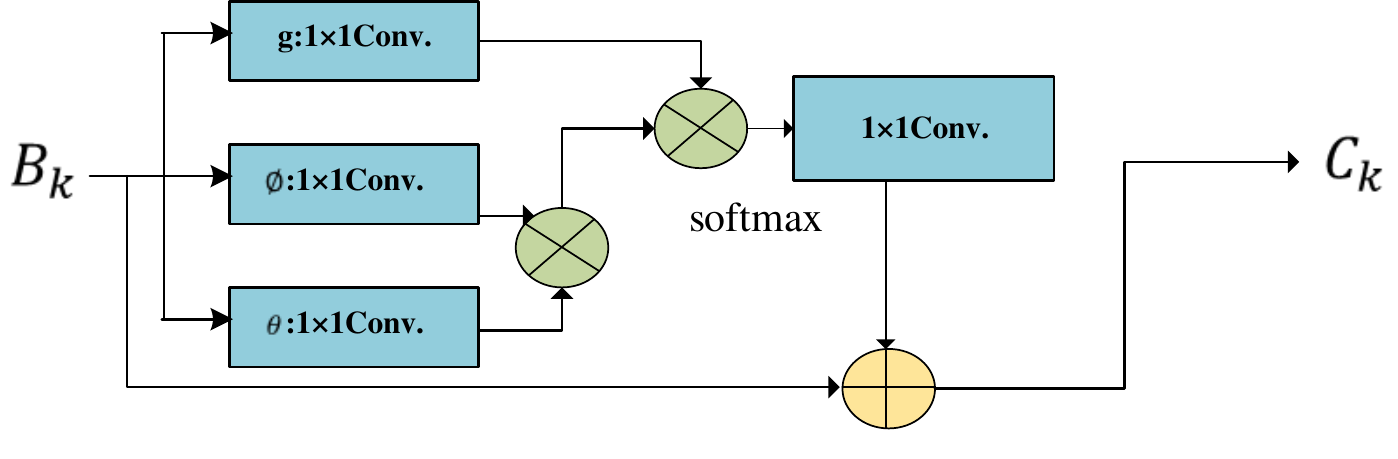}
\caption{{\bf Architecture of Non-Local-Block Layer.} The softmax operation is performed on each row.}
\label{fig:NonLocalBlock}
\end{figure}

\subsection{NLD}
NLD is a SOD model to learn discriminant saliency features and generate saliency map $O(z)$ from the enhanced image $J(z)$. As illustrated in Fig.~\ref{fig:framework}, NLD follows the architecture of our previous work~\cite{xu2018extended}. Different from~\cite{xu2018extended} that conduct low illumination SOD directly on original degraded images, NLD detects salient object from the enhanced images. The Non-Local-Block Layer is utilized to capture the difference of each feature against its local neighborhood favoring regions in the enhanced image. Those regions are either brighter or darker than their neighbors, and the differences catch more details. Therefore, the extracted non-local feature can reflect both the local and global context of an image by incorporating the details of various resolutions. And the detail information of the salient object can be retained to improve the SOD performance.

Fig.~\ref{fig:NLD} illustrates the architecture of NLD subnet for SOD. The first row of NLD contains five convolutional blocks derived from VGG-16 (CONV-1 to CONV-5). The goal of these convolutional layers is to learn feature maps $X_{1}$-$X_{5}$. The second layer contains five convolutional blocks (CONV-6 to CONV-10). Each block changes the number of channels to 128. The goal of these convolutional layers is to learn multi-scale local feature maps $B_{1}$-$B_{5}$. Then, Non-Local-Block Layer obtains more useful features from enhanced images and learns feature maps $C_{1}$-$C_{5}$. The last row is a set of deconvolution layers (UNPOOL-2 to UNPOOL-5) to generate $U_{2}$-$U_{5}$. A 1$\times$1 convolution is added after $C_{1}$ to sum the number of channels to 640, and then the local feature map is obtained. Finally, the SCORE block has 2 convolution layers and a softmax to compute the saliency probability by fusing the local and global features.

As illustrated in Fig.~\ref{fig:NonLocalBlock}, the proposed Non-Local-Block Layer consists of two operations, 1$\times$1 convolution, and softmax. The 1$\times$1 convolution is used to generate feature maps, while the softmax is utilized to store the similarity of any two pixels. Motivated by \cite{Wang2017Non}, the similarity of any two pixels is calculated by non-local mean \cite{buades2005non} and bilateral filters \cite{tomasi1998bilateral}, ensuring the feature map can be embedded into Gaussian after 1$\times$1 convolution. It is formulated as:

\begin{equation}
f\left(x_{i}, x_{j}\right)=e^{\left(W_{\theta} x_{i}\right)^{T} W_{\phi} x_{j}},
\end{equation}
where $x_{i}$ and $x_{j}$ represent two pixels of each feature map $B_{1}$-$B_{5}$. $W_{\theta}$ and $W_{\phi}$ are the weights of the convolution layers. A pairwise function $f$ computes a scalar (representing relationship such as affinity) between $i$ and all $j$. After the convolution, the number of channels becomes a half of the initial size.

The similarity calculated above is stored in the feature maps by the mean of self-attention. It is defined by $y_{k} = softmax(B_{k}^{T} W_{\theta}^{T} W_{\phi} B_{k})g(B_{k})$. For simplicity, we only consider $g$ in the form of a linear embedding: $g\left(B_{k}\right)=W_{g} B_{k}$, where $W_{g}$ is a weight matrix to be learned. Then, we use 1$\times$1 convolutions to recover the number of channels. After that, the feature map $C_{k}$, $k$ = 1,...,5 is obtained through a residual operation using $y_{k}$ and $B_{k}$ via:

\begin{equation}
C_{k}=W_{B} y_{k}+B_{k},
\end{equation}
where $W_{B}$ is a weighting parameter to restore the same number of channels $y_{k}$ as $B_{k}$. ``+$B_{k}$" denotes a residual connection. The residual connection allows us to insert a new non-local block into any pre-trained model. After processing by the non-local network layer $B_{k}$, the size of the feature map $C_{k}$ remains the same. By doing so, the pixel information of feature maps can be reserved.

\subsection{Overall Loss Function}
In PIE, the atmospheric light and transmission map are learned simultaneously, where a joint loss function is utilized to combine the atmospheric light estimation error and the transmission map estimation error. Different from DCPDN~\cite{zhang2018densely} which adopts the $L2$ loss in predicting the atmospheric light, PIE minimizes the MSE loss function between the estimated value $A(z)$ and corresponding ground truth obtained from dark channel prior~\cite{He2009Single}. The MSE loss can be calculated as follows:

\begin{equation}
L^{a}=\frac{1}{N H W} \sum_{i=1}^{N}\left\|A(z)-A_{g t}\right\|^{2},
\end{equation}
where $H$ and $W$ are the height and the width of the image, respectively. And $N$ is the total number of training batches.

In NLD, the local features $L$ and global features $G$ are linearly combined as follows:

\begin{equation}
\hat{y}(v)=p(y(v)=c)=\frac{e^{W_{L}^{c} L(v)+b_{L}^{c}+W_{G}^{c} G+b_{G}^{c}}}{\sum_{c^{\prime} \in\{0,1\}} e^{W_{L}^{c^{\prime}} L(v)+b_{L}^{c^{\prime}}+W_{G}^{c^{\prime}} G+b_{G}^{c^{\prime}}}},
\end{equation}
where ($W_{L}$, $b_{L}$) and ($W_{G}$, $b_{G}$) are two linear operators. $y(v)$ represents the ground truth. The final saliency map is denoted as $\hat{y}\left(v_{i}\right)$.

The cross-entropy loss function can be formulated as follows:

\begin{equation}
H_{j}(y(v), \hat{y}(v))=-\frac{1}{N} \sum_{i=1}^{N} \sum_{c \in\{0,1\}}\left(y\left(v_{i}\right)=c\right)(\log (\hat{y}(v_i)=c))
\end{equation}

In order to make the boundary robust to background noise, the IoU boundary loss of NLDF~\cite{luo2017non} is utilized and can be calculated as follows.

\begin{equation}
\operatorname{IoU}\left(C_{i}, C_{j}\right)=1-\frac{2\left|C_{i} \cap C_{j}\right|}{\left|C_{i}\right|+\left|C_{j}\right|}
\end{equation}

Finally, the overall loss function can be obtained by the combination of the cross-entropy loss function and the IoU boundary loss.

\begin{equation}
\text {Total Loss} \approx \sum_{j} \lambda_{j} \int H_{j}(y(v), \hat{y}(v))+\sum_{j} \gamma_{j}\left(1-\operatorname{IoU}\left(C_{i}, C_{j}\right)\right)
\end{equation}

\section{Nighttime Image Dataset for SOD}
\begin{figure}[h]
\centering
\includegraphics[width=0.8\columnwidth]{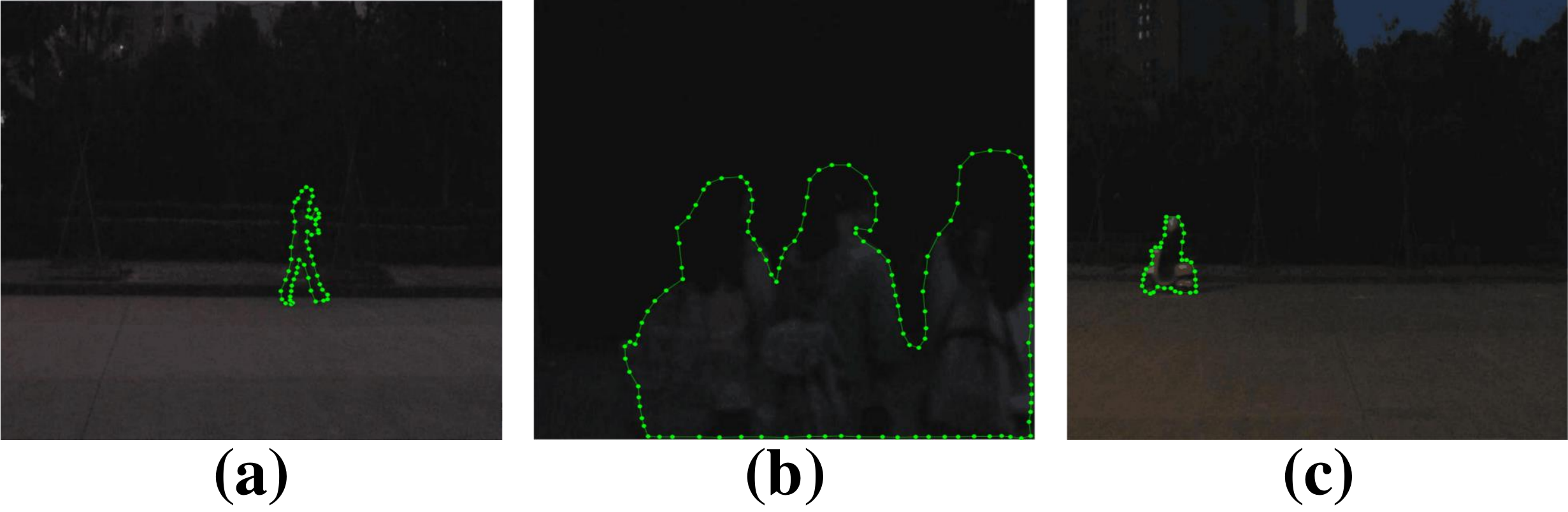}
\caption{{\bf Example samples of the NTI-V1 dataset.}}
\label{fig:example}
\end{figure}

\begin{figure}[h]
\centering
\includegraphics[width=0.6\columnwidth]{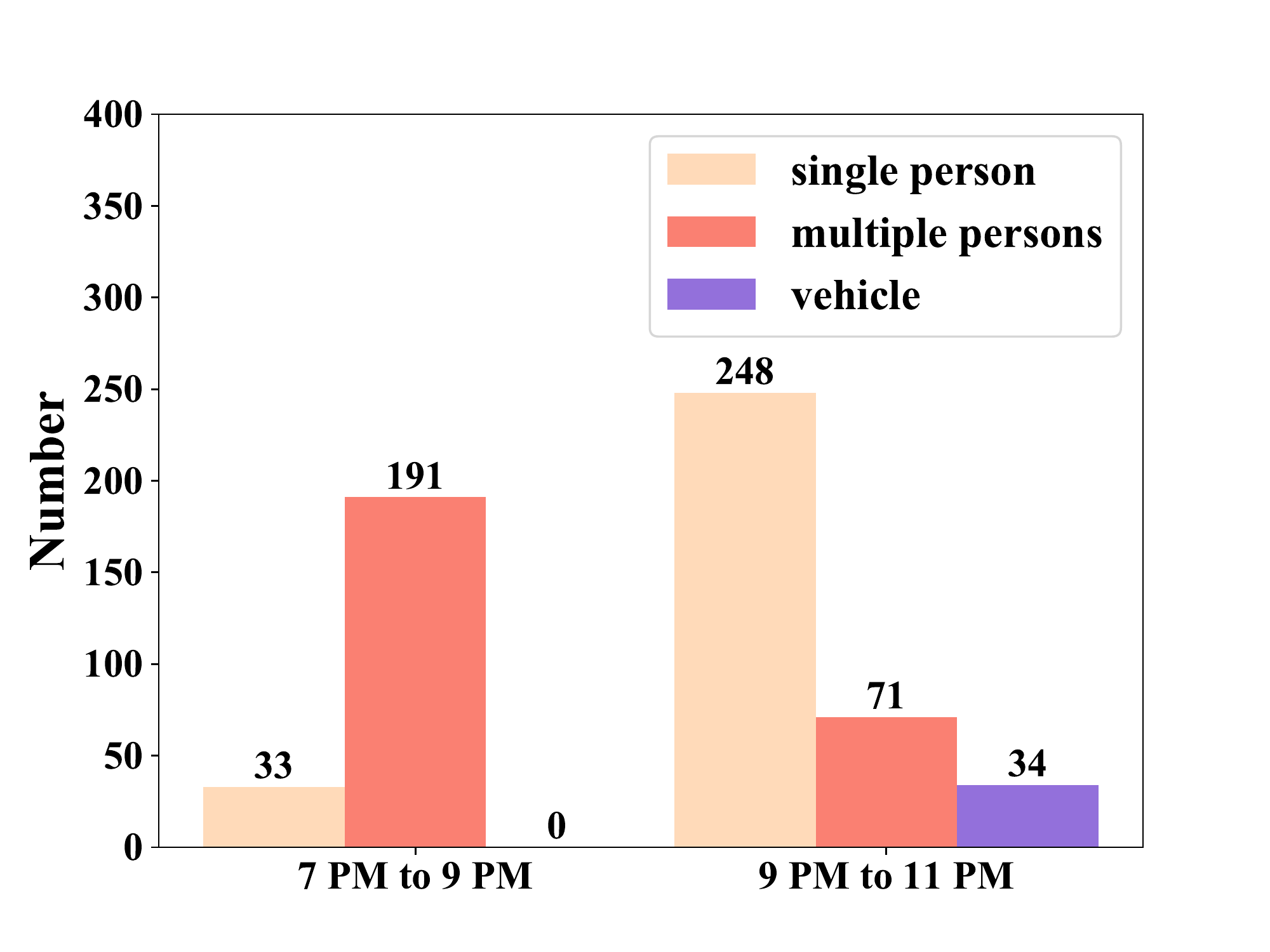}
\caption{\bf The data collection division of the NTI-V1 dataset.}
\label{fig:data_division}
\end{figure}

We build a NightTime Image - V1 (NTI-V1) dataset for SOD. NTI-V1 contains 577 low light images, each image is accompanied by pixel-level human-labeled ground-truth annotation. These images are captured at the nighttime of spring-summer and the autumn-winter from the indoor and the outdoor scene of our university. And we incorporate various challenges, such as viewpoint variation, changing illumination, and diverse scenes. The dataset was collected in two stages. In the first stage, 224 high-resolution images were captured by one surveillance camera from 7 PM to 9 PM. In the second stage, 353 images were captured by three smartphones from 9 PM to 11 PM. After the collection, 5 volunteers are invited to annotate the salient objects with bounding boxes. The shared image regions (with IoU > 0.8) of these bounding boxes are kept as the salient objects. To provide high-quality annotations, we further manually label the accurate silhouettes of the salient objects via the `LabelMe' software. Fig.~\ref{fig:example} shows some examples of the NTI-V1 dataset. The dataset includes 3 types of objects: single person (Fig.~\ref{fig:example}(a)), multiple persons (Fig.~\ref{fig:example}(b)), and vehicle (such as bicycle, car, \textit{etc.}) (Fig.~\ref{fig:example}(c)). Fig.~\ref{fig:data_division} shows the data collection division of the NTI-V1 dataset. In our evaluation protocol, 457 images are used for training, and 120 images are used for testing.

To the best of our knowledge, although there are some datasets for the low light image enhancement, there is no related dataset for evaluating the performance of low light image SOD. In Table~\ref{t:dataset}, we make a comparison of related datasets, including See-In-the-Dark (SID)~\cite{chen2018learning}, LOw Light image dataset (LOL)~\cite{wei2018deep}, SOD~\cite{martin2001database}, MSRA-B~\cite{liu2010learning}, ECSSD~\cite{yan2013hierarchical}, DUT-OMRON~\cite{yang2013saliency}, and PASCAL-S~\cite{li2014secrets}. For the former two datasets, they do not contain salient object segmentation thus inappropriate for SOD. Similar to the following five datasets, each image of the NTI-V1 dataset is accompanied with pixel-level ground-truth annotation. However, those datasets are generally constructed at daytime, containing very few low light images. Hence, our dataset is the first available benchmark dataset for the low light image SOD.

To facilitate the research of low light SOD, we collect a dataset called NTI-V1 Dataset with following distinct features: 1) It contains 577 images captured in nighttime, and each of which is accompanied by pixel-level human-labeled ground-truth annotation. 2) The dataset is captured by one surveillance camera from 7 PM to 9 PM and three smartphones from 9 PM to 11 PM, which covers a large area of districts and at different times. 3) It contains multiple salient objects per image including 3 types of objects: single person, multiple persons, and vehicle (such as bicycle and car). And 4) the capture conditions involve various viewpoints, illumination changes, and different scenes.

\begin{table}
  \caption{{\bf Comparing the NTI-V1 dataset with existing low light image datasets and SOD datasets.}}
  \label{t:dataset}
  \begin{tabular}{l|cccc c}
    \toprule
    Dataset   &\#Img.   &\#DT   &\#NT   &\#Obj. &Annotation \\
    \midrule
    SID       & 5,518   &424    &5,094  &--     &\ding{55}       \\
    LOL       & 1,000   &500    &500    &--     &\ding{55}       \\
    \midrule
    MSRA-B    & 5,000   &4,964  &36      &1-2    &\checkmark      \\
   SOD       & 300     &299    &1      &1-4+   &\checkmark      \\
   ECSSD     & 1,000   &998    &2      &1-4+   &\checkmark       \\
   DUT-OMRON & 5,168   &5,166  &2      &1-4+   &\checkmark       \\
   PASCAL-S  & 850     &842    &8      &1-4+   &\checkmark      \\
   NTI-V1 (Ours)       & 577     &0      &577    &1-4+   &\checkmark         \\
  \bottomrule
\end{tabular}
\end{table}

\section{Experiments}
\subsection{Datasets and Experimental Settings}
We conduct extensive experiments on five SOD datasets, including DUT-OMRON, ECSSD, PASCAL-S, SOD, and our proposed NTI-V1 dataset. The former four are generally built in a bright environment and are widely used in the SOD field. While the latter NTI-V1 dataset is built in the nighttime scene.

\textbf{DUT-OMRON.} The DUT-OMRON dataset consists of 5,168 high-quality images. Images in this dataset have one or more salient objects and relatively complex backgrounds. Thus, it is challenging for saliency detection.

\textbf{ECSSD.} The ECSSD dataset contains 1,000 images with semantic meaning in their ground truth segmentation. It also contains images with complex structures.

\textbf{PASCAL-S.} The PASCAL-S dataset contains 850 challenging images (each composed of several objects), all of which are chosen from the validation set of the PASCAL VOC 2010 segmentation dataset.

\textbf{SOD.} The SOD dataset contains 300 images designed for image segmentation. This dataset provides ground truth for the boundaries of salient objects perceived by humans in natural images.

\textbf{NTI-V1.} We constructed the NTI-V1 dataset which contains 577 natural scene images under low illumination. This dataset contains three types of objects hand-labeled as the ground truth, including single person, multiple persons, and vehicle (such as bicycle, car, and \textit{etc.}).

\textbf{Saliency Evaluation Metrics.} We adopt three widely used metrics to measure the performance of all algorithms, the Precision-Recall (PR) curves, F-measure and Mean Absolute Error (MAE)~\cite{borji2015salient}. The precision and recall are computed by thresholding the predicted saliency map and comparing the binary map with the ground truth. The PR curve of a dataset indicates the mean precision and recall of saliency maps at different thresholds. The F-measure is a balanced mean of average precision and average recall, and is calculated by $F_{\beta}=\frac{\left(1+\beta^{2}\right) \times \text {Precision} \times \text {Recall}}{\beta^{2} \times \text {Precision}+\text {Recall}}$, where $\beta^{2}$ is set to 0.3 to emphasize the precision over recall~\cite{wang2018salient}. The maximum $F_{\beta}$ (max $F_{\beta}$) of each dataset is reported in this paper. We also calculate the MAE for fair comparisons as suggested by ~\cite{borji2015salient}. The MAE evaluates the saliency detection accuracy by MAE$=\frac{1}{W \times H} \sum_{x=1}^{W} \sum_{y=1}^{H}|S(x, y)-L(x, y)|$, where $S(x, y)$ is the predicted salient map and $L(x, y)$ is the ground truth. The parameters $W$ and $H$ represent the width and height of the image, respectively.

\textbf{Implementation Details.} Our model was built on PyTorch. We set related hyper-parameters of PIE following~\cite{zhang2018densely}. During training, we utilized the Adam optimization with the learning rate of $2 \times 10^{-3}$ for both generator and discriminator. All the training samples were resized to 512$\times$512. For the NLD, the weights of CONV-1 to CONV-5 were initialized by the VGG-16 network. All weights in the network were initialized randomly by a truncated normal distribution ($\sigma$ = 0.01) and the biases were initialized to zero. We also used Adam optimization with the learning rate of $10^{-6}$. 457 images from the NTI-V1 dataset were fed into the network for training, and in turn, the other 120 images were used for testing.

\begin{table}
  \caption{{\bf Quantitative results of our method on three types of objects in two stages}.}
  \label{t:NTI}
  \begin{tabular}{l|c|c|c}
    \toprule
    Type   & Criteria & 7 PM - 9 PM    & 9 PM - 11 PM   \\
    \midrule
    \multirow{2}{*}{single person} & MAE$\downarrow$         &0.004 &0.024 \\
                           & max $F_{\beta}\uparrow$ &0.921 &0.695 \\
    \hline
    \multirow{2}{*}{multiple persons}     & MAE$\downarrow$         &0.011 &0.035 \\
                           & max $F_{\beta}\uparrow$ &0.795 &0.706 \\
    \hline
    \multirow{2}{*}{vehicle}  & MAE$\downarrow$         &- &0.018 \\
                           & max $F_{\beta}\uparrow$ &- &0.797 \\
  \bottomrule
\end{tabular}
\end{table}

\subsection{Comparison with State-of-the-art Methods}

To evaluate the proposed algorithm, extensive tests have been performed using a set of SOD methods, including MR~\cite{yang2013saliency}, BSCA~\cite{qin2015saliency}, DSSC~\cite{hou2017deeply}, LPS~\cite{zeng2018learning}, R3Net~\cite{deng2018r3net}, and BASNet~\cite{qin2019basnet}. Qualitative and quantitative evaluations are explored to comprehensively the performance of PIE for SOD with other seven low light image enhancement methods, including  Gamma~\cite{farid2001blind}, LIME~\cite{guo2017lime}, LECARM~\cite{ren2018lecarm}, Dong~\cite{dong2011fast}, Ying~\cite{ying2017new}, RetinexNet~\cite{wei2018deep}, and GLADNet~\cite{wang2018gladnet}.

To investigate the influence of low illumination and object type, we conducted experiments on three types of objects in two stages separately. Tab.~\ref{t:NTI} summaries the results in terms of MAE and max $F_{\beta}$. We can observe that the results of 7 PM to 9 PM is superior to the results of 9 PM to 11 PM, due to the illumination conditions around 7 PM to 9 PM is better. On the other hand, there is no obvious phenomenon for different types of objects.

To evaluate the effectiveness of our image enhancement methods for SOD as well as to promote further research on this new problem, we adopt three types of performance evaluation. (1) To verify the effectiveness of our proposed methods on low light images, we compare our proposed method with several SOD methods on low light images. (2) To verify the effectiveness of our PIE for low light SOD, we enhance low light images by PIE, and then compare our proposed method with several SOD methods on the enhanced images. (3) To verify the appropriateness of PIE for NLD, we compare our proposed method on enhanced images generated via different image enhancement methods.

\subsubsection{Comparison with state-of-the-art SOD methods on Low light images}
\begin{figure*}[t]
	\centering
	\footnotesize{
		\begin{tabular}{cc}
			\includegraphics[width=0.5\textwidth]{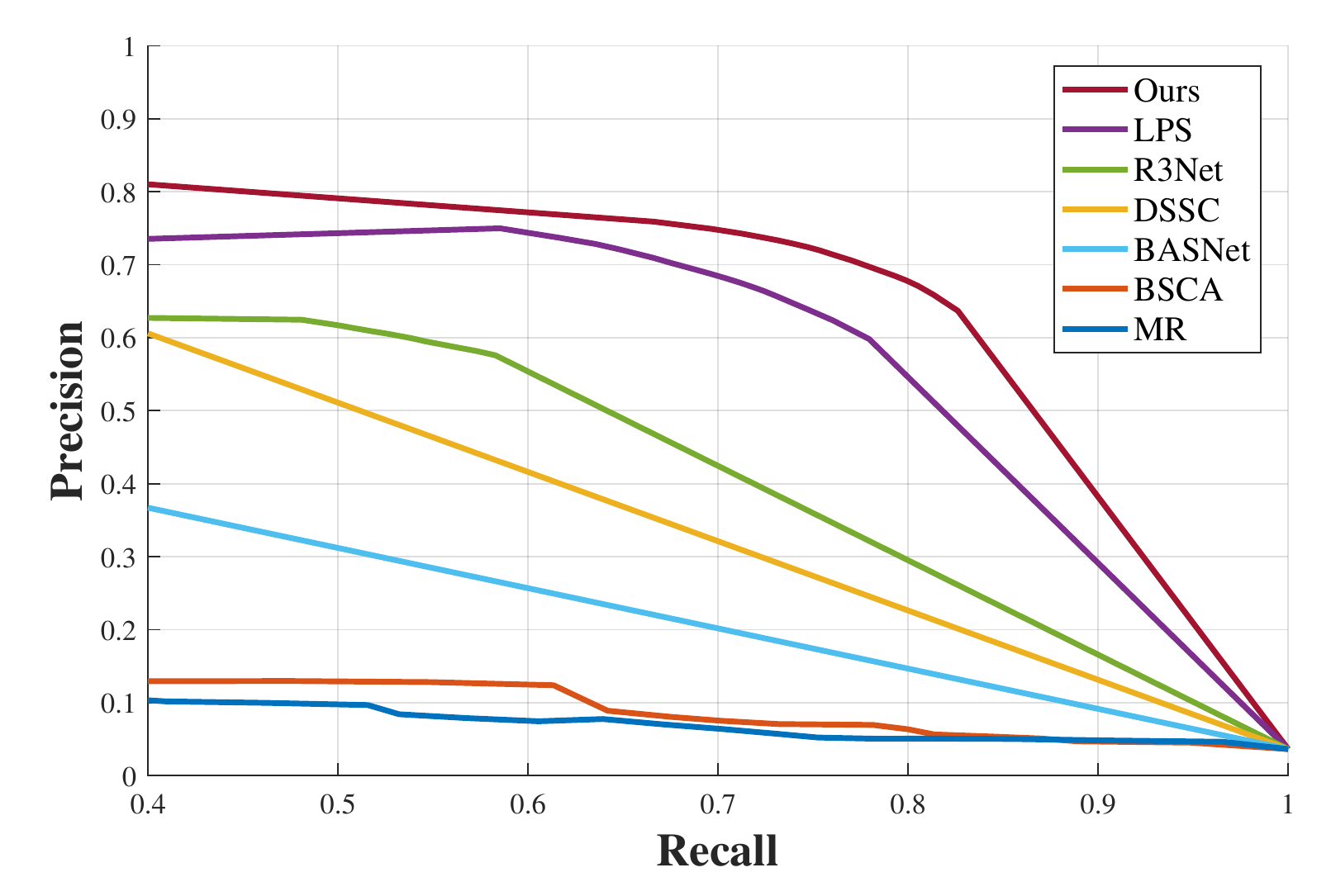} &
			\includegraphics[width=0.46\textwidth]{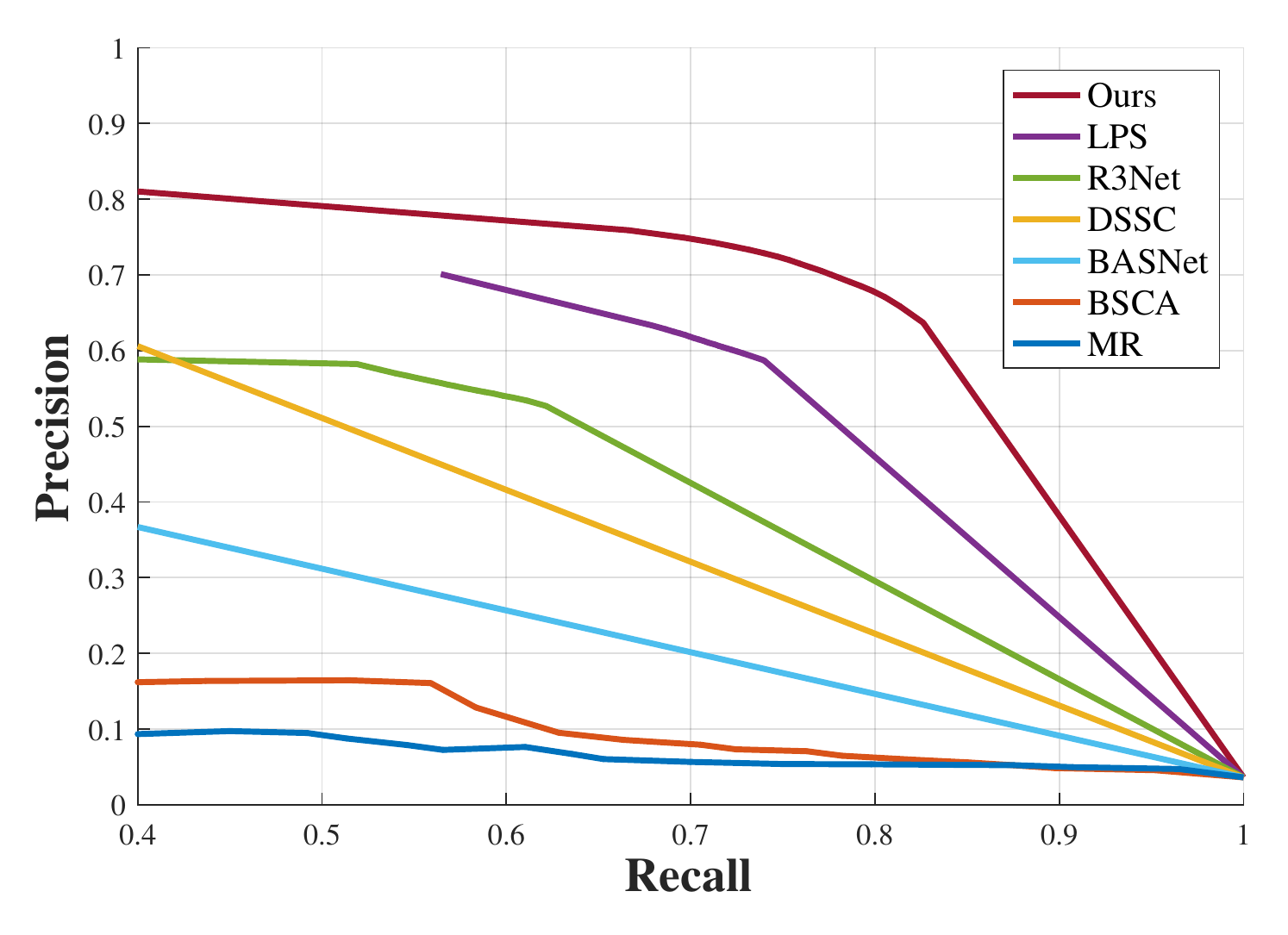}\\
			\small{(a)} & \small{(b)}\\

	\end{tabular}}
	\caption{{\bf PR curves on the NTI-V1 dataset.} (a) PR curves for our method compared to LPS~\cite{zeng2018learning}, R3Net~\cite{deng2018r3net}, DSSC~\cite{hou2017deeply}, BASNet~\cite{qin2019basnet}, BSCA~\cite{qin2015saliency}, and MR~\cite{yang2013saliency} on NTI-V1 dataset. (b) PR curves for our method compared to LPS~\cite{zeng2018learning}, R3Net~\cite{deng2018r3net}, DSSC~\cite{hou2017deeply}, BASNet~\cite{qin2019basnet}, BSCA~\cite{qin2015saliency}, and MR~\cite{yang2013saliency} on enhanced NTI-V1 dataset via PIE.}
	\label{fig:PRC} 
\end{figure*}

We compared our method with several state-of-the-art SOD methods, including MR~\cite{yang2013saliency}, BSCA~\cite{qin2015saliency}, DSSC~\cite{hou2017deeply}, LPS~\cite{zeng2018learning}, R3Net~\cite{deng2018r3net}, and BASNet~\cite{qin2019basnet} on five datasets. Tab.~\ref{t:STOA} shows the comparison results in terms of MAE and max $F_{\beta}$ for all datasets. We can observe that our method does not obtain the best performances on the four public daytime SOD datasets. While our method beats the state-of-the-art methods on the NTI-V1 dataset. To further evaluate the quality of SOD methods, we compared their PR curves on the NTI-V1 dataset, as shown in Fig.~\ref{fig:PRC}(a). Our method achieves a better PR curve than all the other methods. It shows that our method achieves the best performance on the NTI-V1 dataset with respect to both two metrics. It also indicates that our method is more effective to detect salient objects in low light images, although not better than the others on the images with sufficient light.

\begin{figure}[h]
\centering
\includegraphics[width=1.0\columnwidth]{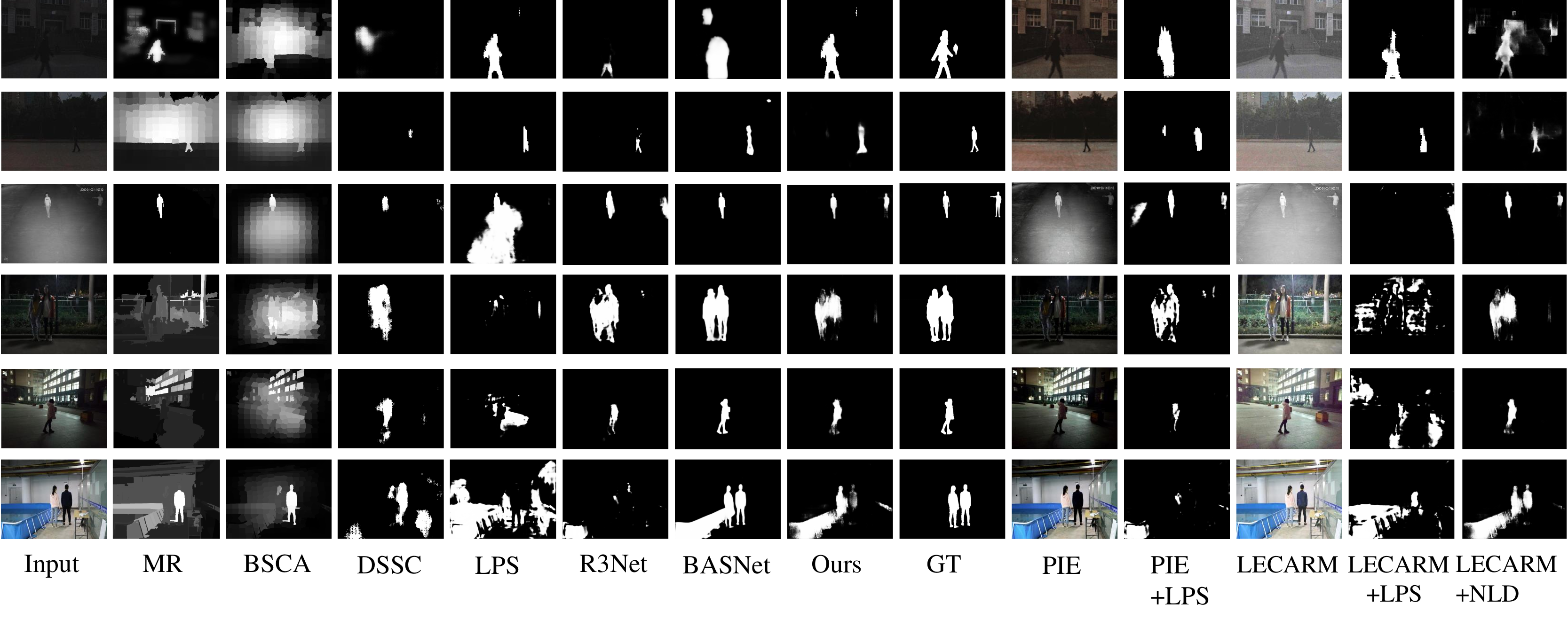}
\caption{{\bf Selected qualitative evaluation results on the NTI-V1 dataset.} The $1st$ column shows the input images. From the $2nd$ column to the $8th$ column are respectively the SOD results of MR~\cite{yang2013saliency}, BSCA~\cite{qin2015saliency}, DSSC~\cite{hou2017deeply}, LPS~\cite{zeng2018learning}, R3Net~\cite{deng2018r3net}, BASNet~\cite{qin2019basnet}, and the proposed method. The $9th$ column is the ground truth. The $10th$ and $12th$ columns are the enhanced images by our PIE and the LECARM method respectively. The $11th$ and $13th$ columns are the SOD results of the LPS method on the enhanced images of the $10th$ and $12th$ columns respectively. The $14th$ column shows the SOD results of our NLD on the enhanced images of the $12th$ column.}
\label{fig:results} 
\end{figure}

In Fig.~\ref{fig:results}, we show the qualitative results. The non-deep learning based methods MR and BSCA perform badly on the low light images. The DSSC method is difficult to learn useful features for the nighttime scene. The pixel-based method LPS produces a lot of false detection due to noise interference. The methods BASNet and R3Net lost many saliency details and tend to contain non-saliency backgrounds. Comparing the results from the $2nd$ to the $8th$ columns, we can observe that our method exhibits sharper and uniformly highlighted salient objects, and the saliency maps are closer to the ground truth (the $9th$ column).

\tabcolsep=6pt

\begin{table}
  \caption{{\bf Benchmarking results of 6 state-of-the-art SOD models on 5 datasets}: DUT-OMRON, ECSSD, PASCAL-S, SOD, and our newly constructed NTI-V1. Three top results are highlighted in \rc{red}, \bc{blue} and \gc{green}, respectively. The up-arrow $\uparrow$ shows the larger value achieves, the better the performance is. The down-arrow $\downarrow$ has the opposite meaning.}
  \label{t:STOA}
  \begin{tabular}{lcccccccc}
    \toprule
    Dataset   & Criteria & MR    & BSCA  & DSSC  & LPS   & R3Net & BASNet & Ours \\
    \midrule
    \multirow{2}{*}{DUT-OMRON} & MAE$\downarrow$         &0.187 &0.191 &0.065 &\gc{0.064} &\bc{0.063} &\rc{0.056} &0.069\\
                           & max $F_{\beta}\uparrow$ &0.610 &0.616 &\gc{0.720} &0.635 &\bc{0.795} &\rc{0.805} &0.667\\
   \midrule
   \multirow{2}{*}{ECSSD}     & MAE$\downarrow$         &0.189 &0.183 &\gc{0.062} &0.087 &\bc{0.040} &\rc{0.037} &0.122\\
                           & max $F_{\beta}\uparrow$ &0.736 &0.758 &\gc{0.873} &0.814 &\bc{0.934} &\rc{0.942} &0.775\\
   \midrule
   \multirow{2}{*}{PASCAL-S}  & MAE$\downarrow$         &0.223 &0.224 &0.103 &\rc{0.041} &\gc{0.092} &\bc{0.076} &0.133\\
                           & max $F_{\beta}\uparrow$ &0.666 &0.666 &\gc{0.773} &0.694 &\bc{0.834} &\rc{0.854} &0.761\\
   \midrule
   \multirow{2}{*}{SOD}       & MAE$\downarrow$         &0.273 &0.266 &0.126 &0.169 &\gc{0.125} &\rc{0.114} &\bc{0.118}\\
                           & max $F_{\beta}\uparrow$ &0.619 &0.634 &0.787 &0.707 &\bc{0.850} &\rc{0.851} &\gc{0.849}\\
   \midrule
   \multirow{2}{*}{NTI-V1}       & MAE$\downarrow$         &0.355 &0.326 &\bc{0.027} &0.029 &0.033 &\gc{0.028} &\rc{0.026}\\
                           & max $F_{\beta}\uparrow$ &0.138 &0.136 &0.481 &\bc{0.678} &\gc{0.591} &0.557 &\rc{0.745}\\
   \bottomrule
\end{tabular}
\end{table}

\begin{table}
  \caption{{\bf Benchmarking results of 6 state-of-the-art SOD methods on enhanced NTI-V1 dataset via PIE.} The dataset was first enhanced by our PIE, then evaluated by different saliency detection models.}
  \label{tab:STOA-2}
  \begin{tabular}{cccccccc}
    \toprule
    Method & MR & BSCA & DSSC & LPS & R3Net & BASNet & \bf{Ours}\\
    \midrule
     MAE$\downarrow$ & 0.351 & 0.306 &\gc{0.029} &0.036  &0.031 &\bc{0.028} & \rc{0.026} \\
    max $F_{\beta}\uparrow$ & 0.140 & 0.144 &0.458 &\bc{0.689} &0.549 &\gc{0.551} & \rc{0.745} \\
  \bottomrule
\end{tabular}
\end{table}

\begin{table}
  \caption{{\bf Benchmarking results of 7 state-of-the-art image enhancement methods on NTI-V1 dataset for SOD by NLD.} The dataset was first enhanced by different image enhancement models, then evaluated with our saliency detection model NLD.}
  \label{tab:STOA-3}
  \begin{tabular}{ccccccccc}
    \toprule
     Method & Gamma & LIME & LECARM & Dong & Ying & RetinexNet & GLADNet & \bf{Ours}\\
    \midrule
     MAE$\downarrow$ & 0.060 & \gc{0.050} & 0.052 & 0.067  & 0.064 & \bc{0.041} & 0.052 & \rc{0.026}\\
     max $F_{\beta}\uparrow$ & \gc{0.575} & 0.484 & \bc{0.577} & 0.469 & 0.494 & 0.424 & 0.465 & \rc{0.745}\\
  \bottomrule
\end{tabular}
\end{table}

\subsubsection{Comparisons with several SOD methods on low light images enhanced via PIE}
To verify the effectiveness of our PIE for low light SOD, we firstly enhanced the NTI-V1 datasets by our PIE, then the enhanced images were trained and tested by our NLD and the other state-of-the-art SOD methods. Tab.~\ref{tab:STOA-2} shows the comparison results in terms of MAE and max $F_{\beta}$ on the NTI-V1 dataset. It is obvious that PIE can improve the performance of SOD compared to Tab.~\ref{t:STOA}. To further evaluate the quality of SOD methods, we compared their PR curves on the NTI-V1 dataset, as shown in Fig.~\ref{fig:PRC}(b). Our method achieves a better PR curve than all the other methods on the NTI-V1 dataset. It shows that our method achieves the best performance on the NTI-V1 dataset with respect to both two metrics. Moreover, our image enhancement method PIE not only have effectiveness for our SOD model (NLD), but also other existing SOD models.

In Fig.~\ref{fig:results}, we show the qualitative results. Comparing the `PIE' and `Input', we can observe that PIE improves the brightness and contrast of the low light images obviously, outstanding the salient object. Furthermore, it is observed that `PIE+LPS' ($11th$ column) achieved better SOD results than the method `LPS' ($5th$ column), which shows the effectiveness of PIE for SOD. Furthermore, from these results, we can obverse that `PIE+LPS' ($11th$ column) include non-saliency backgrounds compared to `Ours' ($8th$ column). Our method ($8th$ column) is more close to the ground truth (the $9th$ column).

\subsubsection{Comparisons among PIE and several image enhancement methods}
To verify the appropriateness of PIE for NLD, low light images were respectively enhanced by Gamma~\cite{farid2001blind}, LIME~\cite{guo2017lime}, LECARM~\cite{ren2018lecarm}, Dong~\cite{dong2011fast}, Ying~\cite{ying2017new}, RetinexNet~\cite{wei2018deep}, GLADNet~\cite{wang2018gladnet}, and our PIE. Then, the enhanced images were trained and tested by our NLD. Tab.~\ref{tab:STOA-3} shows the comparison results in terms of MAE and max $F_{\beta}$ on NTI-V1 dataset. It is clear that our method achieves the best results with respect to both two metrics, verifying the appropriateness of PIE for NLD.

In the Fig.~\ref{fig:results}, we show the qualitative results. The `PIE' ($10th$ column) and `LECARM' ($12th$ column) can improve the brightness and contrast of the low light images obviously compared with the `Input' ($1st$ column). Furthermore, the results of `PIE+LPS' ($11th$ column) and `LECARM+LPS' ($13th$ column), `Ours'($8th$ column) and `LECARM+NLD' ($14th$ column) indicate that the PIE's performance improvement for LPS lower than LECARM's, but the PIE's performance improvement for NLD better than LECARM's. Visualization results also show this phenomenon that our method achieved better SOD results than the methods `PIE+LPS' ($11th$ column) and `LECARM+NLD' ($14th$ column). From these results, we can obverse that `PIE+LPS' ($11th$ column) and `LECARM+NLD' ($14th$ column) tend to lose many saliency details and include non-saliency backgrounds. Our method can accurately segment salient objects in low light.

\subsection{Ablation Study}
The ablation experiments are conducted on the NTI-V1 dataset.
\subsubsection{The Ablation Study of $A(z)$.} Our PIE differs from DCPDN~\cite{zhang2018densely}. We believe that the atmospheric light $A(z)$ is a point-wise random value rather than a constant. So we treat $A(z)$ as a random value instead of a constant. Here, we compared our method with a constant $A(z)$. Tab.~\ref{t:ablation2} shows that our method improves MAE by 1.4\% and max $F_{\beta}$ by 0.4\% on the NTI-V1 dataset. It validates our designing for the atmospheric light $A(z)$.

\subsubsection{The Ablation Study of Non-Local-Block Layer.}
The structure of NLD is similar to NLDF~\cite{luo2017non}. However, our NLD utilizes additional Non-Local-Block (NLB) Layers to calculate the similarity among different pixels. Tab.~\ref{t:ablation2} shows that our method improves MAE by 2.1\% and max $F_{\beta}$ by 4.1\% on the NTI-V1 dataset, compared with our method without the NLB Layer. It validates our designing for the NLB Layer.

\begin{table}
  \caption{{\bf Ablation study results of $A(z)$ and Non-Local-Block Layer on the NTI-V1 dataset.} Note that our method is with a random value $A(z)$ and the NLB Layer.}
  \label{t:ablation2}
  \begin{tabular}{c|cc}
    \toprule
    Method         & MAE$\downarrow$ & max $F_{\beta}\uparrow$ \\
    \midrule
    Ours w/ Constant $A(z)$ & 0.040      & 0.741   \\
    Ours w/o NLB Layer  & 0.047      & 0.704      \\
    {\bf Ours}          & {\bf 0.026}  & {\bf 0.745} \\
  \bottomrule
\end{tabular}
\end{table}

\section{Conclusion}
In this work, we propose an image enhancement based SOD for low light images, which is critical for CV applications in low light conditions~\cite{zeng2020illumination,kansal2020sdl}. This method directly embeds the physical lighting model into the deep neural network to describe the degradation of low light images, and in turn, utilizes a Non-Local-Block Layer to extract non-local features of salient objects. Further, we construct an NTI-V1 dataset containing 577 low light images with pixel-wise object-level annotations for the SOD community. With extensive experiments, we verify that low illumination can actually reduce the performance of SOD.
The proposed method is effective to enhance local content in low light images to facilitate the SOD task.

\section{Acknowledgments}
This work was supported by the Natural Science Foundation of China (U1803262, 61602349, and 61440016).


\bibliographystyle{ACM-Reference-Format}
\bibliography{sample-base}

\appendix

\end{document}